\documentclass{article}
\usepackage{ijcai25}

\usepackage{times}
\usepackage{soul}
\usepackage{url}
\usepackage[hidelinks]{hyperref}
\usepackage[utf8]{inputenc}
\usepackage[small]{caption}
\usepackage{graphicx}
\usepackage{amsmath}
\usepackage{amsthm}
\usepackage{booktabs}
\usepackage{algorithm}
\usepackage[switch]{lineno}

\usepackage{textcomp}
\usepackage{enumitem}
\usepackage{tikz}

\usepackage{textcomp}
\usepackage{stfloats}
\usepackage{url}
\usepackage{verbatim}
\usepackage{graphicx}
\usepackage{relsize}

\usepackage{xcolor}
\usepackage{algorithm}
\usepackage{algpseudocode}
\usepackage{booktabs,arydshln}
\usepackage{multirow}
\usepackage{enumitem}
\usepackage{tikz}
\usepackage{threeparttable}
\usepackage{url}
\usepackage{amssymb}

\makeatletter
\newcommand\plainfootnote[1]{%
  \begingroup
  \renewcommand\thefootnote{}\footnotetext{#1}%
  \addtocounter{footnote}{-1}%
  \endgroup
}
\makeatother

\title{FedAGHN: Personalized Federated Learning with Attentive Graph HyperNetworks}

\author{
Jiarui Song$^1$\thanks{Equal contribution.}
\and
Yunheng Shen$^1$$^*$\and
Chengbin Hou$^{2}$\thanks{Corresponding Authors.}\and\\
Pengyu Wang$^{3}$\and 
Jinbao Wang$^{4}$\and 
Ke Tang$^{5}$\And
Hairong Lv$^1$$^\dagger$\\
\affiliations
$^1$Tsinghua University
$^2$Fuyao University of Science and Technology\\
$^3$Hong Kong University of Science and Technology
$^4$Shenzhen University\\
$^5$Southern University of Science and Technology\\
}
\begin{document}

%$^\ddagger$ 
%$^\dagger$

\maketitle

\plainfootnote{This is the authors' final accepted manuscript of the article published in \textit{Knowledge-Based Systems}, 2025. The published version is available at: \url{https://doi.org/10.1016/j.knosys.2025.114355}}

\plainfootnote{
© 2025. This manuscript version is made available under the CC BY-NC-ND 4.0 license \url{https://creativecommons.org/licenses/by-nc-nd/4.0/}
}

\begin{abstract}
Personalized Federated Learning (PFL) aims to address the statistical heterogeneity of data across clients by learning the personalized model for each client. Among various PFL approaches, the personalized aggregation-based approach conducts parameter aggregation in the server-side aggregation phase to generate personalized models, and focuses on learning appropriate collaborative relationships among clients for aggregation. However, the collaborative relationships vary in different scenarios and even at different stages of the FL process. To this end, we propose Personalized Federated Learning with Attentive Graph HyperNetworks (FedAGHN), which employs Attentive Graph HyperNetworks (AGHNs) to dynamically capture fine-grained collaborative relationships and generate client-specific personalized initial models. Specifically, AGHNs empower graphs to explicitly model the client-specific collaborative relationships, construct collaboration graphs, and introduce tunable attentive mechanism to derive the collaboration weights, so that the personalized initial models can be obtained by aggregating parameters over the collaboration graphs. Extensive experiments can demonstrate the superiority of FedAGHN. Moreover, a series of visualizations are presented to explore the effectiveness of learned collaboration graphs.
\end{abstract}

\section{Introduction}
Data privacy and security concerns in the training of deep learning models, along with the phenomenon of isolated data islands, have attracted significant attention from researchers, government agencies, and other stakeholders.
Federated Learning (FL) has provided a new paradigm for training the global model through client-side collaboration in distributed systems without the sharing of clients' private data. The idea behind it is to transmit model parameters rather than private local data. For instance, FedAvg \cite{pmlr-v54-mcmahan17a} allows clients to perform local training and then upload the update of local model parameters to the server for average aggregation, resulting in a global model that is subsequently distributed back to the clients.
% However, the data distributions among different clients exhibit statistical heterogeneity in real-world scenarios, posing challenges for traditional FL approaches which rely on a single global model to adapt for the diverse local data of different clients \cite{9492755mtl}.
However, in real-world scenarios, the data distributions among different clients exhibit statistical heterogeneity, posing significant challenges for traditional FL approaches, which rely on a single global model, in adapting to the diverse local data of different clients \cite{9492755mtl}.

Personalized Federated Learning (PFL) offers a novel perspective to address statistical heterogeneity. The main concept is for each client to learn a personalized model, rather than sharing a single global model.
It enables each client to learn a personalized local model by acquiring information from other clients and/or the global model. 
There exist various approaches to PFL, such as local adaptation, meta-learning-based \cite{metalearningjiang2019improving,metalearningfallah2020personalized}, multi-task-learning-based \cite{multitaskmarfoq2021federated}, regularization-based \cite{NEURIPS2020_f4f1f13c,pmlr-v139-li21h}, parameter decoupling \cite{pmlr-v139-collins21a,chen2022onFedRoD,10.1145/3580305.3599345FedCP}, knowledge distillation-based \cite{wu2022communication,xu2023personalized} and personalized aggregation-based approaches \cite{li2021fedphp,Zhang_Hua_Wang_Song_Xue_Ma_Guan_2023,he2024privacy}. 
Among these, the personalized aggregation-based approach naturally fits the need for personalized adaptation to the local data of different clients. Herein, client-specific personalized models are generated based on collaborative relationships among the clients. In this work, we further investigate this approach with a focus on the collaboration mechanism and collaborative relationship learning.

In the personalized aggregation-based approach, personalized aggregation is performed in the server-side aggregation phase instead of the global average aggregation. The aim is mainly to obtain appropriate collaborative relationships for parameter aggregation, such that each client can receive additional gains without sharing local data \cite{ge2024fedaga}. To achieve the goal, three key points may need to be considered:
\begin{enumerate}[leftmargin=*]
    \item \textit{Collaboration weights and weight differences.} The collaboration weights offer a pattern for aggregating clients' parameters and producing the personalized aggregated model. The weight differences of different FL scenarios should be adapted to the specific FL scenario considering its data distributions or other characteristics over the clients.
    \item \textit{Self-participation strength in collaboration.} The self-participation strength refers to the self-weighting of client's personalized local model during the personalized aggregating phase. %Each client should determine its inclination towards retaining its own learned information versus the information received from other collaborating clients.
    Each client should evaluate its preference for retaining its learned information.
    \item \textit{Layer-wise collaboration differences.} For the deep learning models with multiple layers, personalized aggregation should consider the layer-wise personalization. Past studies \cite{lecun2015deep,Ma_2022_CVPR,Lee_Zhang_Avestimehr_2023LA} have suggested that different layers have different behaviours, e.g., the shallow layers capture generic information, whereas the deep layers acquire specific information.
\end{enumerate}
\vspace{1pt}

%bengio2013representation

\begin{figure}[!t]
\centering
\includegraphics[width=0.48\textwidth]{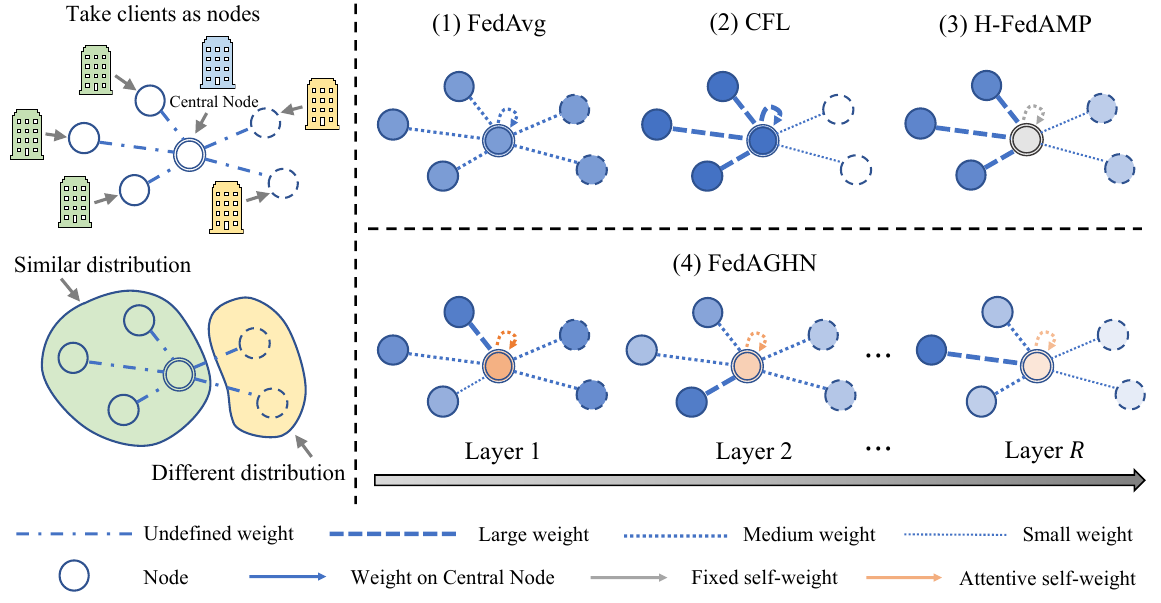}
\caption{Comparison of previous works and our approach in modeling collaborative relationships. A darker node color indicates a higher collaborative weight of the corresponding edge. 
Our FedAGHN can adaptively tune the attentive weights among clients for layer-wise personalized aggregation.}
\label{fig_intro}
\end{figure}

%and continuously 
With regard to the three aforementioned points, most existing methods focus only on certain aspect(s).~For example, as shown in~Figure~\ref{fig_intro}, CFL \cite{9174890CFL} only considers collaboration with clustered intra-group collaborating clients; H-FedAMP \cite{Huang_Chu_Zhou_Wang_Liu_Pei_Zhang_2021} considers the differences among collaborating clients, but the collaboration weight of the central client remains fixed. pFedLA \cite{Ma_2022_CVPR}, FedALA \cite{Zhang_Hua_Wang_Song_Xue_Ma_Guan_2023} and FedLGS \cite{he2024privacy} optimize trainable parameters to implicitly capture relationships, but encounter challenges in effectively capturing appropriate collaborative relationships.

To fulfill all three key points, we propose a novel personalized aggregation-based FL method, \textit{Personalized \underline{Fed}erated Learning with \underline{A}ttentive \underline{G}raph \underline{H}yper\underline{N}etworks (FedAGHN)}, which employs the proposed Attentive Graph HyperNetworks (AGHNs) to generate client-specific personalized initial models. In each communication round, AGHNs maintain a set of collaboration graphs for each client and each layer. The collaboration graph is initialized by taking both personalized local model parameters and its updates as node features. A tunable attentive mechanism is introduced to calculate collaboration weights, leveraging prior knowledge from the updates of models and two trainable parameters optimized towards the client's objective. One trainable parameter is employed to explicitly tune the weight differences of collaborating clients (Point 1), and another one is used to adjust the self-participation strength in the collaboration (Point 2). After that,
the personalized initial model of a client at each layer (Point 3) can be obtained by weighted aggregation over the corresponding learned graph. These points are also the main innovations of our FedAGHN.

The main contributions of this work are as follows:
\begin{itemize}[leftmargin=*]
    \item We propose a novel personalized aggregation-based method called FedAGHN, which naturally integrates graph learning into the FL process. FedAGHN considers weight differences, self-participation strength, and layer-wise personalization to better capture the fine-grained collaboration weights for personalized aggregation. 
    %This may offer a graph perspective for the future FL research.
    % to explicitly consider weight differences and self-participation strength
    \item We introduce two sets of trainable parameters in FedAGHN, which are beneficial for capturing the dynamic evolution of collaborative relationships. These trainable parameters are optimized dynamically in the FL process to align with the objective of each client for learning the client-specific personalized model.
    \item We conduct extensive experiments and discover several meaningful insights by visualizing and analyzing the collaboration graphs from different points of view, such as overall patterns, layer-wise patterns, and the evolution of patterns during the FL process. The code is available at \url{https://github.com/songjiarui00/FedAGHN2025}.
\end{itemize}

\section{Related Work}

\subsection{Personalized Federated Learning}

Statistical heterogeneity has become a crucial challenge in FL settings \cite{ding2024global,zhang2025personalized,zhang2024structural}.

PFL aims to learn models that better fit each client's local data, enabling them to effectively tackle the statistical heterogeneity in FL \cite{zheng2025personalized,9599369survey}. Recently, the development of PFL has led to the emergence of various approaches \cite{9743558TowardsPFL,zhang2023pfllib}.
Vanilla PFL methods, such as FedAvg-FT and FedProx-FT, employ local training of the global model to obtain personalized models, which can be classified as \textit{local adaptation methods}. 
On the other hand, \textit{meta-learning-based methods} \cite{metalearningjiang2019improving,metalearningfallah2020personalized} and \textit{multi-task-learning-based methods} \cite{multitaskmarfoq2021federated} apply the principles of meta-learning and multi-task learning, respectivety, to PFL, aiming to enhance the model adaptation.

\textit{Regularization-based methods}, such as pFedMe \cite{NEURIPS2020_f4f1f13c} and Ditto \cite{pmlr-v139-li21h}, have new ideas for using global model in PFL. These methods learn personalized models for each client and employ L2 regularization to constrain the distance between the personalized models and the global model. 
Another feasible way to absorb model information is by knowledge distillation. FedPAC \cite{xu2023personalized} is one of a number of \textit{knowledge distillation-based methods}, which introduce global feature centroids for explicit local-global feature representation alignment. 

\textit{Parameter decoupling methods} are commonly employed to achieve better personalized or shared models. 
For example, FedRep \cite{pmlr-v139-collins21a} first trains personalized classification layers and then trains a feature extractor, FedCP \cite{10.1145/3580305.3599345FedCP} handles global and personalized information in the features separately using a global head and a personalized head, employing conditional policies to differentiate between global and personalized information. FedAS \cite{yang2024fedas} uses parameter alignment and client synchronization mechanisms to suppress inconsistencies between the personalized parameter part and the shared parameter part.

\textit{Personalized aggregation-based methods} aim to obtain relationships among clients and leverage them as a basis for
aggregation. Specifically, CFL \cite{9174890CFL} classifies clients into groups, with each group sharing the same personalized model aggregated within the group. 
FedAMP and H-FedAMP \cite{Huang_Chu_Zhou_Wang_Liu_Pei_Zhang_2021} address personalized aggregation with a client-wise approach using attention-based message passing mechanisms. pFedLA \cite{Ma_2022_CVPR} produces layer-wise aggregation weights and implicitly optimizes them via hypernetworks, whereas FedALA \cite{Zhang_Hua_Wang_Song_Xue_Ma_Guan_2023} learns collaboration matrices between local models and the global model for aggregation. FedLGS \cite{he2024privacy} uses a multi-layer neural network to learn the aggregated weights of different clients.

Earlier-developed \textit{personalized aggregation-based methods} focus mainly on certain aspects of collaboration among clients. By contrast, our work uses collaboration graphs to model the fine-grained collaborative relationships, and the collaboration weights on the graphs can then be adaptively optimized with clients’ objectives during the FL process.

\begin{table*}[ht]
\centering
\small
\begin{tabular}{p{2cm} p{4cm} p{4cm} p{4cm}}
\toprule
\textbf{Hypernetworks} & \textbf{FedHN} & \textbf{pFedLA} & \textbf{FedAGHN} \\
\midrule
\textbf{Number} 
& One HN shared by all clients 
& One HN maintained per client 
& One HN maintained per client \\
\midrule
\textbf{Personalization} 
& Client-wise
& Layer-wise
& Layer-wise \\
\midrule
\textbf{Architecture} 
& Neural network (large parameter size) 
& Neural network (moderate parameter size) 
& Lightweight graph neural network \\
\midrule
\textbf{Optimization} 
& Optimizing HN with client-specific embeddings 
& Optimizing client-specific HN with embeddings 
& Optimizing key tunable parameters in HN \\
\midrule
\textbf{Output}
& Directly generating personalized model parameters 
& Generating layer-wise aggregation coefficients for personalized model generation 
& First constructing collaboration graphs; then generating personalized initial model parameters \\
\bottomrule
\end{tabular}
\caption{Comparison of Hypernetwork-based Methods}
\label{tab:hn_comparison}
\end{table*}

\subsection{Graph Learning}
Graph learning has attracted much attention in recent years \cite{10.1145FSGNN}. The attention mechanism is introduced over the relational weights to be optimized with certain objectives \cite{velivckovic2017gat,zhang2024label}, such that the optimized weights better reflect the relations between nodes in terms of the objectives. In PFL, each client could have different specific relations with other clients because of inherent different data distributions, and oftentimes, the relations between the clients are often unknown and complicated. It thus becomes critical to learn the relations between clients, so as to facilitate the model aggregation (i.e., collaboration) step and boost the performance of the local model.

The collaboration graph can be used to represent the relationship between clients in FL \cite{zhang2022graph}. 
However, natural graph structures do not exist in many real-world scenarios, thus requiring graph structure learning on the server side to establish collaboration graphs among clients \cite{FGML10.1145/3575637.3575644}. A common approach is to construct collaboration graphs based on the model parameters of different clients. For example, SFL \cite{chen2022personalized} constructs a similarity matrix using the similarity of personalized model parameters, then it forms a graph with trainable parameters as edge weights, and optimizes the graph with regularization during local training at the client side; whereas pFedGraph \cite{pmlr-v202-ye23b} designs optimization goals on the server side and proceeds to optimize the collaboration graph by solving a quadratic program. FedPnP \cite{rasti2025fedpnp} constructs graph relationships between clients based on label distribution and performs personalized collaboration according to the graph relationships.

Existing approaches thus far treat the model as a whole entity when constructing and utilizing the collaboration graph, disregarding the differences between different model layers. However, it is crucial to learn specific collaboration graphs for each model layer. By contrast, our work learns collaboration graphs for different model layers, these graphs can then be optimized via a tunable attentive mechanism during the FL process.

\subsection{Hypernetworks}

The hypernetwork technique utilizes neural network to generate parameters (such as weights) for the target network, with the objective of enhancing its performance in target tasks.
%Hypernetworks have been extensively applied in machine learning tasks such as computer vision \cite{klocek2019hypernetwork}, language modeling \cite{NIPS2017_f9d11525HN}, neural architecture search (NAS)~\cite{zhang2018graphHN}.
This property of hypernetworks, enabling the generation of improved client model parameters in the target task, makes them highly suitable for PFL \cite{10.1093/nsr/nwae132}. For example, pFedHN \cite{pmlr-v139-shamsian21a} maintains a unique embedding vector for each client, serving as input to the hypernetwork for generating personalized model parameters. By optimizing both the embedding vectors and hypernetwork parameters, pFedHN facilitates the generation of improved personalized model parameters. HAFL-GHN \cite{litany2022federated} focuses on generating distinct model architecture weights for different clients. PLFL \cite{wang2025personalized} proposes a lightweight hypernetwork to generate client model parameters and uses pruning strategies for personalized personalization. Meanwhile, rather than directly generating the complete weights of models, pFedLA \cite{Ma_2022_CVPR} utilizes hypernetworks to generate layer-wise weighted aggregation parameters. pFedLA has been verified to yield hypernetwork-generated aggregation parameters that can capture fine-grained relationships among clients, thereby enhancing the performance of personalized models. Additionally, the hypernetworks are on the server side, which mitigates the introduction of additional communication overhead.

The hypernetwork architecture in pFedLA employs SoftMax layers to generate the weights with differences for different model layers. On the other hand, in pFedHN, the majority of hypernetwork parameters are shared by all clients. The optimization process involves a larger number of parameters, which may potentially result in smaller differences between clients or between layers. 

Regarding our work, we reduce the number of trainable parameters in the hypernetwork and retain only the vital training parameters that can tune the collaboration graphs, thus reducing the optimization burden while increasing the differences among model layers.
A clearer comparison of the similarities and differences between our FedAGHN and these hypernetwork-based PFL methods is shown in the Table \ref{tab:hn_comparison}.

\section{The Proposed Method}

In this section, we first formally formulate the problem of PFL. After that, we present the overall framework and algorithm of the proposed FedAGHN. Finally, we elaborate on the technical details of AGHNs step by step. 

\subsection{Problem Formulation}
The objective of PFL is to learn a personalized local model for each client without sharing their private data. If it is assumed that there are $N$ clients, and that $\Theta=\left \{ \theta_1,\dots,\theta_N \right \}$ represents the parameters of these personalized local models, trained on their local datasets $\mathcal{D}=\{\mathcal{D}_1,\dots,\mathcal{D}_N\}$, where the datasets $\mathcal{D}$ are sampled from Non-IID settings. The objective function of PFL can be formulated as
\begin{equation}
\Theta^*=\arg\min_{\Theta}\frac{1}{N}\sum_{i=1}^{N} \sum_{(x,y) \sim \mathcal{D}_i} \mathcal{L}_i\left ( \theta_i;x,y  \right ) 
\end{equation}
% \begin{equation}
% \Theta^*=\arg\min_{\Theta}\frac{1}{N}\sum_{i=1}^{N}\frac{1}{|\mathcal{D}_i|} \sum_{(x,y) \sim \mathcal{D}_i} \mathcal{L}_i\left ( \theta_i;x,y  \right ) 
% \end{equation}
% where 
% \begin{equation}
% \mathcal{L}_i\left (\theta_i  \right ) = \frac{1}{|\mathcal{D}_i|} \sum_{(x,y) \sim \mathcal{D}_i} \mathcal{L}\left ( \theta_i;x,y  \right ) 
% \end{equation}
%and $\mathcal{L}_i$ is the empirical loss function of client $i$ on dataset $\mathcal{D}_i$. 
where $\mathcal{L}_i$ is the empirical loss of client $i$ on dataset $\mathcal{D}_i$. 
% Besides, the cross-entropy loss could be adopted for $\mathcal{L}$ and $|\mathcal{D}_i|$ counts the number of samples in $\mathcal{D}_i$. 

Inherited from traditional federated learning, PFL also contains two main phases: local update and aggregation communication.
For most PFL methods, the vanilla local update phase of the $t$-th round can be written as
\begin{equation}
\label{local_update_formulate} 
    \theta_i^{(t)} = \theta_i^{(t)} - \eta \nabla_{\theta_i} \mathcal{L}_i(\theta_i^{(t)})
\end{equation}
where $\eta$ is the client learning rate and $\theta_i^{(t)}$ denotes the personalized local model of the $i$-th client in the $t$-th communication round.

Then in the aggregation phase, the server collects the updates of personalized local model parameters from each client and aggregates them to obtain a global model by 
\begin{equation}
\label{aggregation_formulate}
    {\theta}_\text{g}^{(t+1)} = \sum_{i=1}^{N} \frac{|\mathcal{D}_i|}{|\mathcal{D}|}\theta_i^{(t)}
\end{equation}
where $\theta_\text{g}^{(t+1)}$ denotes the aggregated global model. This global model can be seen as having absorbed diverse information from heterogeneous clients in a sense.

After that, each client could use the aggregated global model as its initialization model in the next local update phase, i.e.,
\begin{equation}
\label{init_formulate}
    \bar{\theta}_1^{(t+1)} = \cdots = \bar{\theta}_N^{(t+1)} = \theta_\text{g}^{(t+1)}
\end{equation}
where $\bar{\theta}_i^{(t+1)}$ denotes the personalized initial model of client $i$ in the next round $t+1$ and set $\theta_i^{(t+1)}=\bar{\theta}_i^{(t+1)}$ in the beginning of the next local update phase.

However, due to the existence of data heterogeneity, the local models updated by Eq. \eqref{local_update_formulate} exhibit significant differences among various clients, which leads to the difficulty in obtaining a globally excellent model on all clients as shown in Eq. \eqref{aggregation_formulate}. Our AGHN method addresses this issue by adopting a personalized aggregation approach, designing a separate aggregation function for each client through graph hypernetworks.

Personalized aggregation-based methods focus mainly on the aggregation step to aggregate personalized models for each client, i.e.,
% where modifying Eq. (\ref{aggregation_formulate}) and Eq. (\ref{init_formulate}) as
\begin{equation}
    \bar{\theta}_i^{(t+1)} = \text{Agg}_i(\Theta^{(t)})
\label{gene-aggfunc}
\end{equation}
where $\bar{\theta}_i^{(t+1)}$ denotes the personalized initial model of client $i$ in the next round $t+1$ and $\text{Agg}_i$ represents the aggregation function for different personalized aggregation-based methods. In our method, $\text{Agg}_i$ is specified as the $\text{AGHN}_i$, as shown in Eq. (\ref{eq:AGHN}).

\subsection{FedAGHN Algorithm}

Given that capturing the better fine-grained collaborative relationships among clients is the most essential and bottom-level requirement in the heterogeneous scenario of PFL, we propose a novel aggregation-based PFL method, namely Personalized Federated Learning with Attentive Graph HyperNetworks (FedAGHN), which utilizes our proposed Attentive Graph HyperNetworks (AGHNs) to generate client-specific personalized initial models.

The proposed FedAGHN aims to provide the collaboration graphs (i.e., collaboration weights) for the model aggregation among $N$ clients. FedAGHN also integrates prior knowledge to explicitly model the collaboration among clients, and it can adaptively adjust collaborative relationships through the optimization of the hypernetworks' parameters with respect to the clients' objectives in an end-to-end fashion.

Specifically, we maintain a dedicated AGHN at the server side for each client, which signifies that there are $\text{AGHNs}=\{\text{AGHN}_1, \text{AGHN}_2, \dots,\text{AGHN}_N\}$ on the server. The AGHNs are employed to capture the collaboration among clients, generating collaboration graphs $G=\{G_1,G_2,\dots,G_N\}$ for each client. Each graph $G_i$ also contains multiple subgraphs at the layer-wise level. Based on these layer-wise collaboration graphs, the AGHNs then produce personalized initial models $\bar{\Theta}=[\bar{\theta}_1,\bar{\theta}_2,\dots,\bar{\theta}_N]$ which are dedicated to each client, rather than aggregating a single global model for all clients. In the server aggregation phase of each round $t$, the personalized initial model $\bar{\theta}^{(t+1)}_i$ in the next round $t+1$ for client $i$ can be obtained by
\begin{equation}
\bar{\theta}^{(t+1)}_i=\text{AGHN}_i(\Theta^{(t)},{\Delta\Theta}^{(t)}; p_i; q_i) \label{eq:AGHN}
\end{equation}
where the input of $\text{AGHN}_i$ has two parts: the collection of personalized local model parameters denoted as ${{\Theta}^{(t)}}$, and the set of the updates of personalized local model parameters returned by the clients denoted as ${\Delta\Theta}^{(t)}$. 
It should be noted that $p_i$ and $q_i$ are the parameters of $\text{AGHN}_i$, which are vital for AGHNs to explicitly model and tune the collaborative relationships.

%\textbf{\mbox{Figure~\ref{fig_fedaghn_framework}}}
The overall framework of FedAGHN is illustrated in Figure~\ref{fig_fedaghn_framework}, and the pseudocode of FedAGHN algorithm is presented in Algorithm \ref{FedAGHN_alg}.
%and its pseudocode is provided in Appendix~B.

\newcommand*{\circled}[1]{\raisebox{.4pt}{\textcircled{\raisebox{-.8pt} {#1}}}}

\begin{figure*}[htbp]
\centering
\includegraphics[width=0.98\textwidth]{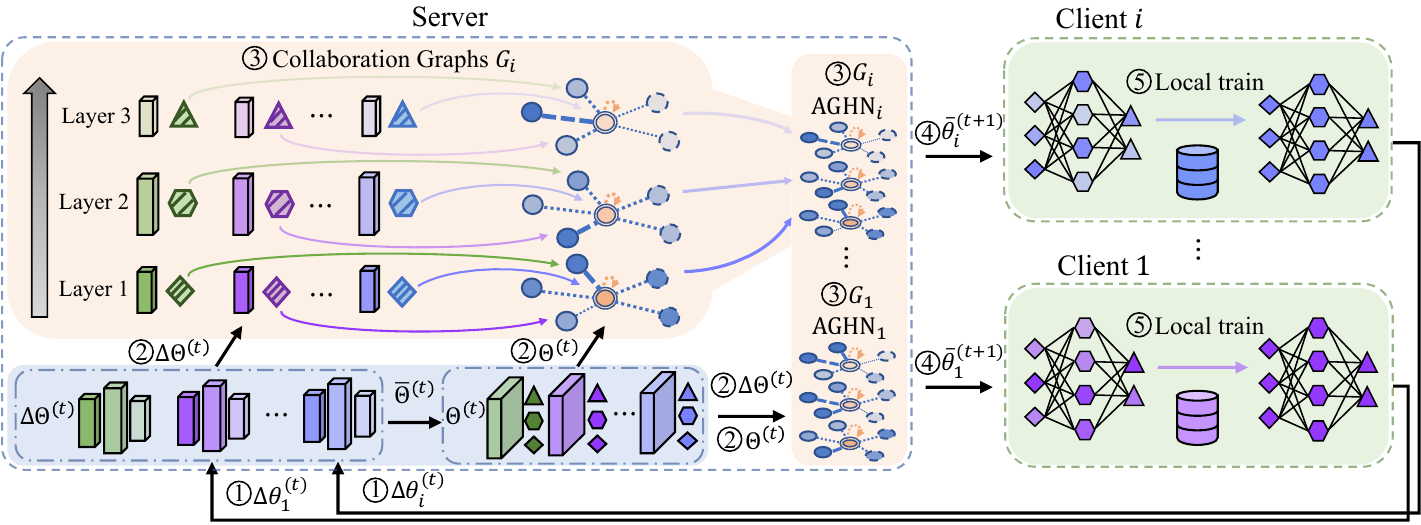}
\caption{The overall framework of FedAGHN. \circled{1} Server side collects $\Delta\Theta^{(t)}=\{\Delta\theta_1^{(t)},\Delta\theta_2^{(t)},\dots,\Delta\theta_N^{(t)}\}$, where the update of personalized local model parameters $\Delta\theta_i^{(t)}$ is sent from client $i$.  \circled{2} $\text{AGHN}_i$ takes $\Delta\Theta^{(t)}$ and $\Theta^{(t)}$ as input, where $\Theta^{(t)}=\bar{\Theta}^{(t)}+\Delta\Theta^{(t)}$ is calculated at server. \circled{3} $\text{AGHN}_i$ updates trainable parameters, generates collaboration graphs $G_i$ with attentive weights, and outputs the personalized initial model $\bar{\theta}_i^{(t+1)}$ as detailed in Section \ref{AGHN}. \circled{4} $\bar{\theta}_i^{(t+1)}$ is sent to client $i$ in order to initialize the local model. \circled{5} Client side executes local training to adapt for the private local data and obtain personalized local model $\theta_i^{(t+1)}$.}
\label{fig_fedaghn_framework}
\end{figure*}

\begin{algorithm}[ht!]
\caption{FedAGHN Algorithm}
\label{FedAGHN_alg}
\textbf{Input}: $N$ clients with local dataset $\mathcal{D}$, global communication rounds $T$, local training epoch $S$, learning rate $\eta$, learning rate of hypernetwork $\eta_{hn}$.\\
\textbf{Parameter}: Initialize the model parameters $\Theta^{(0)}$ and the update of model parameters $\Delta\Theta^{(0)}$ with the same random parameters, initialize AGHNs' trainable parameters $\mathbf{p}$ and $\mathbf{q}$.\\
\textbf{Output}: Trained personalized local models for each client $\Theta = \{{\theta}_1, {\theta}_2, \ldots, {\theta}_N\}$.\
\begin{algorithmic}[1]
\Procedure{\textbf{Sever Executes}}{}
  \For{communication round $t \in \{0, \ldots, T-1\}$}
    \For{client $i$ in parallel}
      \State $\bar{\theta}^{(t+1)}_i = {\text{AGHN}}_i(\Theta^{(t)}, \Delta\Theta^{(t)} ; p_i ; q_i)$
      \State $\theta^{(t+1)}_i,\Delta\theta^{(t+1)}_i = \text{ClientUpdate}(\bar{\theta}^{(t+1)}_i)$
      \State $p_i := \max(0,p_i - \eta_{hn} (\nabla_{p_i} \bar{\theta}^{(t+1)}_i)^{\text{T}} \Delta \theta^{(t+1)}_i)$
      \State $q_i := q_i - \eta_{hn} (\nabla_{q_i} \bar{\theta}^{(t+1)}_i)^{\text{T}} \Delta \theta^{(t+1)}_i$
    \algnotext{EndFor}
    \EndFor
  \EndFor
  \State \textbf{return} $\Theta$
  \algnotext{EndProcedure}
\EndProcedure
\Procedure{\textbf{ClientUpdate}} {$\bar{\theta}_i$}
  \State Client $i$ receives $\bar{\theta}_i$ from the server.
  \State Set $\theta_i = \bar{\theta}_i$.
  \For{local epoch $s \in \{0, \ldots, S-1\}$}
    \For{mini-batch $b_t \subseteq D_i$}
      \State $\theta_i := \theta_i - \eta \nabla_{\theta_i} \mathcal{L}_i(\theta_i; b_t)$
    \EndFor
  \EndFor
  \State $\Delta\theta_i = \theta_i - \bar{\theta}_i$
  \State \textbf{return} $\theta_i, \Delta\theta_i$
\EndProcedure

\end{algorithmic}
\end{algorithm}

\begin{figure}[ht!]
\centering
%origin 2.5in
\includegraphics[width=0.48\textwidth]{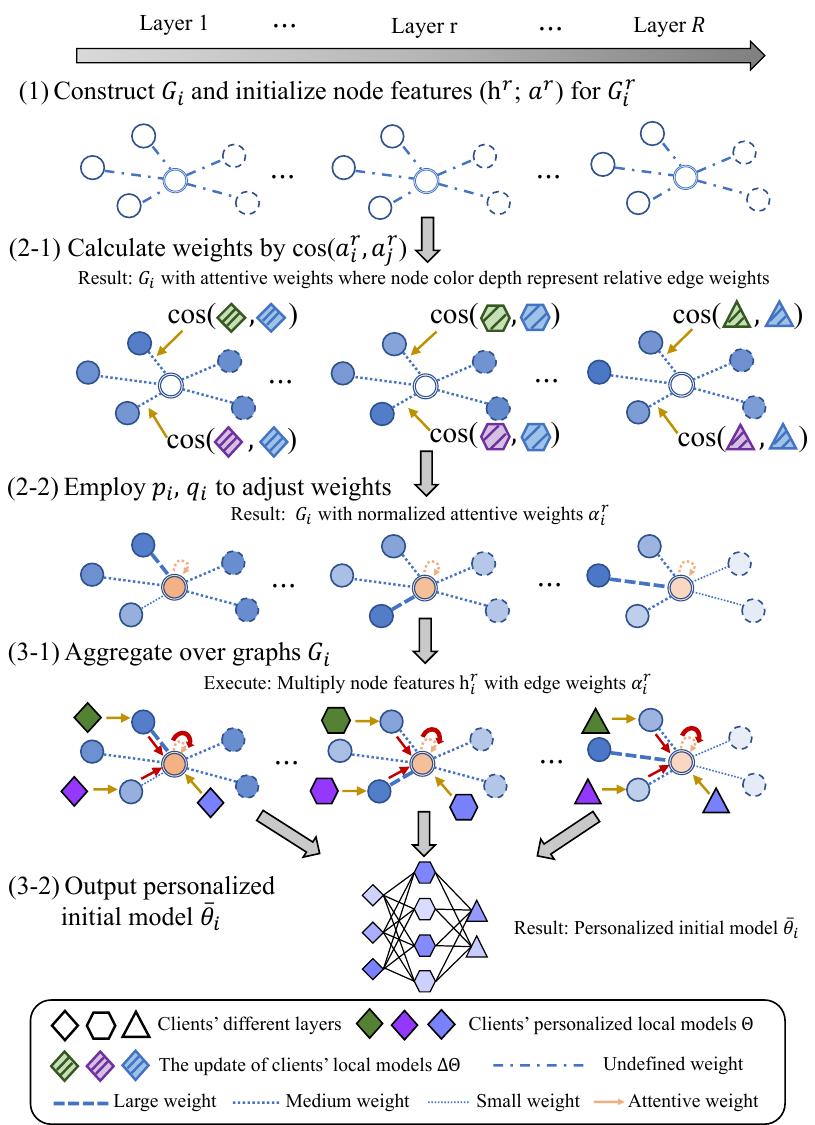}
\caption{The illustration of the attentive graph hypernetwork at client $i$. $\text{AGHN}_i$ takes $\Theta^{(t)}$ and ${\Delta\Theta}^{(t)}$ as inputs and sequentially performs the following steps: (1) initializing collaboration graphs, (2) calculating attentive weights, and (3) aggregating parameters over graphs. Finally, $\text{AGHN}_i$ outputs the personalized initialized model $\bar{\theta}^{(t+1)}_i$.}
\label{fig_aghn}
\end{figure}

\subsection{Attentive Graph Hyper-Networks}\label{AGHN}
In this section, we take the attentive graph hypernetwork for client $i$ (i.e., $\text{AGHN}_i$) as an example to elaborate the technical details. For simplicity, we now focus on one communication round and remove the superscript reflecting the round. 
The AGHN and its workings are illustrated in Figure~\ref{fig_aghn}. The internal computation of AGHN in a detailed pseudocode format is also shown in Algorithm \ref{AGHN_alg}.

{
\begin{algorithm}[ht!]
\caption{Attentive Graph Hyper-Networks}
\label{AGHN_alg}
\textbf{Input}: The model parameters $\Theta^{(t)}$ and the update of model parameters $\Delta\Theta^{(t)}$ in the round $t$.\\
\textbf{Parameter}: ${\text{AGHN}}_i$'s trainable parameters $p_i$ and $q_i$.\\
\textbf{Output}: The personalized initial model $\bar{\theta}^{(t+1)}_i$ of client $i$ for the next round $t+1$.\
\begin{algorithmic}[1]
\Procedure{${\text{AGHN}}_i$} {$\Theta^{(t)}, \Delta\Theta^{(t)} ; p_i ; q_i$}
  \State Set $h=\Theta^{(t)}$ and $a=\Delta\Theta^{(t)}$.
  \For{each layer $r \in \{1, \ldots, R\}$}
    \State Initialize the collaboration graph $G_i^r$ 's topology.
    \State Initialize of node features $(h^r; a^r)$:
    \State $h^r=\{h_1^r,h_2^r,...,h_N^r\}=\{\theta_1^r,\theta_2^r,...,\theta_N^r\}$
    \State $a^r=\{a_1^r,a_2^r,...,a_N^r\}=\{\Delta\theta_1^r,\Delta\theta_2^r,...,\Delta\theta_N^r\}$.
    \State Calculate the collaboration weights:
    \State $\widetilde{\alpha}_{ij}^r= \left\{\begin{matrix}
 \frac{ {e^{q_i^r \text{cos}(a_i^r,a_j^r)}}}{ {\textstyle \sum_{j} e^{q_i^r \text{cos}(a_i^r,a_j^r)} } } & \quad j \ne i\\
 {p_i^r} & \quad j = i
\end{matrix}\right.$.
    \State Normalize the attentive weights:
    \State $\alpha_{ij}^r = \frac{\widetilde{\alpha}_{ij}^r}{{\textstyle \sum_{j=1}^{n}}\widetilde{\alpha}_{ij}^r} \quad \forall j\in\{1,2,\dots,N\}.$
    \State Aggregate parameters over graphs:
    \State ${\bar{\theta}}_i^r = \sum_{j \in \{1,2,\dots,N\}} \alpha_{ij}^r {h_j^r}.$

  \EndFor
  
  \State Collect $\bar{\theta}^{(t+1)}_i = \{ {\bar{\theta}}_i^1, {\bar{\theta}}_i^2, \dots, {\bar{\theta}}_i^R \}$.
  
  \State \textbf{return} $\bar{\theta}^{(t+1)}_i$
\EndProcedure

\end{algorithmic}
\end{algorithm}
}

%(without accessing original data)\cite{lecun2015deep,Ma_2022_CVPR,Lee_Zhang_Avestimehr_2023LA}
The key principle of AGHNs is to learn the collaboration graphs (i.e., collaboration weights) such that each client can aggregate its beneficial model parameters  from all clients based on the learned collaboration graphs. Because the layers of deep learning models may have different functionalities, we also consider different collaboration patterns for different layers respectively, such that $\text{AGHN}_i$ for client $i$ maintains $R$ collaboration graphs $G_i=\{G^{1}_{i},G^{2}_{i},\dots,G^{R}_{i} \}$ where $R$ is the number of layers in the model at client $i$. It should be noted that there is no collaboration among the different layers.

For convenience, we follow the graph learning terminology in describing the AGHN at client $i$ and layer $r$, which corresponds to the collaboration graph $G_i^r$.

\subsubsection{Initialize collaboration graphs}

For the initialization of the graph topology, the collaboration graph $G^r_i$ is established for the central node (client) $i$, such that there are totally $N$ nodes with edges connected to node $i$ from all nodes $j$ (including self-connection from $i$). 

For the initialization of node features $(h^r;a^r)$, we take the personalized local model parameters $\Theta^r$ from clients at layer $r$ to obtain the set of node features $h^r$ as
% \begin{equation}\label{eq:model_para}
% h^r=\{h_1^r,h_2^r,...,h_N^r\}=\{\theta_1^r,\theta_2^r,...,\theta_N^r\}
% \end{equation}
$h^r=\{h_1^r,h_2^r,\dots,h_N^r\}=\{\theta_1^r,\theta_2^r,\dots,\theta_N^r\}$
and we take the updates of personalized local model parameters $\Delta\Theta^r$ from clients at layer $r$ to obtain the set of node features $a^r$ as
% \begin{equation}\label{eq:model_update_para}
% a^r=\{a_1^r,a_2^r,...,a_N^r\}=\{\Delta\theta_1^r,\Delta\theta_2^r,...,\Delta\theta_N^r\}
% \end{equation}
$a^r=\{a_1^r,a_2^r,\dots,a_N^r\}=\{\Delta\theta_1^r,\Delta\theta_2^r,\dots,\Delta\theta_N^r\}$
where the set of node features $h^r$ is employed for aggregation over graphs, whereas the set of node features $a^r$ is utilized to calculate attentive weights. The initial node features for the first round are randomly parameterized, whereas the initial node features for each of the other rounds come from the immediately preceding round.

\subsubsection{Calculate attentive weights}\label{sec:calweight}
Given an initialized collaboration graph $G_i^r$, the node features $a^r$, containing the update of personalized local model parameters $\Delta\theta^r$ from the previous round, can be regarded as prior knowledge. 

Specifically, we calculate the collaboration weights between node $i$ and each node $j\in\{1,2,\dots,N\}$ via a newly proposed tunable attentive mechanism:
\begin{equation}\label{eq:unnorm_att_weight}
{
\widetilde{\alpha}_{ij}^r= \left\{\begin{matrix}
 \frac{ {e^{q_i^r \text{cos}(a_i^r,a_j^r)}}}{ {\textstyle \sum_{j} e^{q_i^r \text{cos}(a_i^r,a_j^r)} } } & \quad j \ne i\\
 {p_i^r} & \quad j = i
\end{matrix}\right.
}
\end{equation}
where $\widetilde{\alpha}_{ij}^r$ denotes the attentive weight between node $i$ and node $j$ at layer $r$, and $\text{cos}(a_i^r,a_j^r)$ gives the prior knowledge based on cosine similarity of the two clients' previous update of personalized local model parameters.
%as defined by Eq. (\ref{eq:model_update_para}). 

To optimize the attentive weights with respect to the client $i$'s objective, we introduce two trainable parameters to further tune the attentive weights. Concretely, $q^r_i$ is adopted to tune the attentive weights between central node $i$ and the other nodes, while $p^r_i$ is used to tune the attentive weight of self-connection of central node $i$. 
Note that, the two trainable parameters $q^r_i$ and $p^r_i$ are vital for $\text{AGHN}_i$\footnote{Remarks of trainable parameters: An increased value of $q_i^r$ denotes that client $i$ prefers greater difference among collaboration weights at layer $r$, while an increased value of $p_i^r$ strengthen the self-weight for client $i$ and thus reduce the inclination to cooperate with other clients. The detailed analysis of $p$ and $q$ can be found in the experiments such as Section \ref{sec:ablation} and \ref{sec:collaboration}.}, because they can explicitly capture and tune the collaborative relationships according to the client's objective in the end-to-end paradigm. 
%The updating rules for $q^r_i$ and $p^r_i$ are detailed in Section \ref{sec:update}.

%\ref{sec:ablation} and

To accelerate convergence and avoid exploding gradients, the normalized attentive weights $\alpha_i^r=\{\alpha_{i1}^r,\alpha_{i2}^r,\dots,\alpha_{iN}^r\}$ for collaboration graph $G_i^r$ can be calculated via 
\begin{equation}\label{eq:att_weight}
\alpha_{ij}^r = \frac{\widetilde{\alpha}_{ij}^r}{{\textstyle \sum_{j=1}^{n}}\widetilde{\alpha}_{ij}^r} \quad \forall j\in\{1,2,\dots,N\}.
\end{equation}

\subsubsection{Aggregate parameters over graphs}\label{sec:aggregate}
The collaboration graph $G_i^r$ includes the attentive weights $\{\alpha_{i1}^r,\alpha_{i2}^r,\dots,\alpha_{iN}^r\}$ and the node features $h^r$ from the previous round. The personalized initial model $\bar{\theta}_i^r$ for client $i$ at layer $r$ can be obtained by weighted aggregating over $G_i^r$ as follows:
\begin{equation}
\label{eq:aghn-aggfunc}
{\bar{\theta}}_i^r = \sum_{j \in \{1,2,\dots,N\}} \alpha_{ij}^r {h_j^r}.
\end{equation}

It might be worth noting that attentive weights are usually calculated using node features $h$ when aggregating $h$ in traditional graph learning, which is different from utilizing another node features $a$ in our implementation described herein.
In fact, this is exactly the unique change we make when applying the graph learning paradigm to federated learning.
This is because the update of model parameters represents the updated direction of the model on local data in FL, it can better reflect the similarity of data distributions among different clients than the model parameters can, thereby better modeling collaborative relationships among the clients.

% (for connections to other clients) (for self-connections)
\subsubsection{The update of AGHN}\label{sec:update}
% Following the previous works \cite{pmlr-v139-shamsian21a,Ma_2022_CVPR}, we accordingly adopt a more general way to update $p_i$ and $q_i$ based on gradient descent, the details can be found in Appendix B.
The collaboration weights among clients should adapt to the changes over communication rounds in FL. For brevity, considering each of the $R$ layers at client $i$, the two sets of trainable parameters are denoted as $q_i=\{q_i^1,q_i^2,\dots,q_i^R\}$ and $p_i=\{p_i^1,p_i^2,\dots,p_i^R\}$, where each layer at client $i$ maintains a pair of scalars $p_i^r$ and $q_i^r$.

Following earlier research studies \cite{pmlr-v139-shamsian21a,Ma_2022_CVPR}, we adopt a more general way to update $p_i$ and $q_i$ based on gradient descent, their updating rules are as follows
\begin{equation}
\label{pq_update}
\begin{split}
q_i &:= q_i - \eta_{hn} \Delta q_i = q_i - \eta_{hn} (\nabla_{q_i} \bar{\theta}_i)^{\text{T}} \Delta \theta_i \\
p_i &:= \max(0,p_i - \eta_{hn} \Delta p_i) \\ & \hspace{0.3em}= \max(0,p_i - \eta_{hn} (\nabla_{p_i} \bar{\theta}_i)^{\text{T}} \Delta \theta_i)
\end{split}
\end{equation}
where $\eta_{hn}$ is the learning rate for the hyper-networks; $\Delta{q_i}$ denotes the gradient vector with $R$ dimensions for client $i$'s updating; $\nabla_{q_i}$ is the partial derivative with respect to $q_i$; the operator $\text{T}$ indicates the transpose operation; $\bar{\theta}_i$ represents the personalized initial model parameters after aggregation; and $\Delta \theta_i$ denotes the update of the personalized local model parameters. It might be worth mentioning that there is no additional communication cost while calculating the trainable parameters $p_i$ and $q_i$.

\section{Experimental Settings}
% This section briefly describes the experimental settings. More detailed experimental settings are presented in Appendix C.

This section introduces the experimental settings, including datasets, models, data heterogeneity, baselines, evaluation, and implementation details. The more specific summary of settings is listed in Table \ref{tab:exp_setup} for better reproducibility.
\begin{table}[ht]
\centering
\small
\begin{tabular}{lll}
\toprule
\textbf{Datasets \& Models} & \textbf{Details} \\
\midrule
 CIFAR-10 & 4-layer CNN + BN, 10 classes \\
 CIFAR-100 & 4-layer CNN + BN, 100 classes \\
 CIFAR-100* &  ResNet-18, 100 classes \\
 Tiny-ImageNet & ResNet-18, 200 classes \\
\midrule
\textbf{Data Heterogeneity} & \textbf{Details} \\
\midrule
Pathological & Randomly assign 5/20/40 classes \\
Practical & Dirichlet distribution $\beta=0.1$ \\
\midrule
\textbf{Training Settings} & \textbf{Details} \\
\midrule
Optimizer & SGD \\
Local Batch Size & 64 \\
Local Epochs & 5 \\
Learning Rate & 0.01 (models), 0.005 (hypernetwork) \\
Client Number & 20 \\
Comm. Rounds & 200/80/40 (empirical convergence) \\
Split Ratio & Train:Val:Test = 7:1:2 per client \\
\bottomrule
\end{tabular}
\caption{Comprehensive Experimental Settings}
\label{tab:exp_setup}
\end{table}

%(denoted as 6-layer CNN), where the 2 BN layers are added after the convolutional layers  for image classification tasks
\subsection{Datasets and Models} 
To evaluate the performance of FedAGHN, we perform a number of tests on three benckmark datasets: CIFAR10, CIFAR100 \cite{krizhevsky2009learning}, and Tiny-ImageNet \cite{chrabaszcz2017downsampled}. For CIFAR10 and CIFAR100, we use the 4-layer CNN architecture with Batch Normalization (BN) layers similar to that done in past research studies \cite{Ma_2022_CVPR}. To assess the performance on larger backbone networks and datasets, we utilize ResNet-18 as the backbone model for CIFAR100 and Tiny-Imagenet. In the following sections, $\text{CIFAR100}^*$ denotes using ResNet-18 on CIFAR100, and TINY indicates using ResNet-18 on Tiny-ImageNet.

\subsection{Data Heterogeneity}
The data heterogeneity is considered through two Non-IID scenarios: the pathological setting and the practical setting. 

In the pathological setting, we adopt the strategy similar to FedAvg \cite{pmlr-v54-mcmahan17a}, where each client is randomly assigned an equal amount of data. Each client is randomly assigned 5, 20, or 40 classes from CIFAR10, CIFAR100 and Tiny-Imagenet, where these datasets contain a total of 10, 100, or 200 classes respectively. 

For the practical setting, we apply the Dirichlet distribution \textit{Dir}($\beta$) with the parameter $\beta$ serving as an indicator of the Non-IID degree. A smaller $\beta$ indicates stronger data heterogeneity. Following prior works \cite{pmlr-v202-ye23b,10.1145/3580305.3599345FedCP}, we set $\beta=0.1$ for all datasets default.

For both settings, we divide the local dataset into a training set, a validation set, and a test set for each client after distributing the whole dataset. The data split ratio for training, validation, and test is set to 7:1:2.

\subsection{Baselines and Evaluation}
The proposed FedAGHN is compared against 15 baselines, including the local training method, 2 traditional federated learning methods, i.e., FedAvg \cite{pmlr-v54-mcmahan17a} and FedProx~\cite{MLSYS2020_1f5fe839}, and 12 personalized federated learning methods. The representative PFL methods cover the variants of FedAvg and FedProx with fine-tuning (local adaptation approach), Ditto \cite{pmlr-v139-li21h} (regularization-based approach),  FedRep \cite{pmlr-v139-collins21a} and FedCP \cite{10.1145/3580305.3599345FedCP} (parameter decoupling approach), pFedHN~\cite{pmlr-v139-shamsian21a} (hypernetwork-based approach), FedPAC \cite{xu2023personalized} (knowledge distillation-based approach), and CFL \cite{9174890CFL}, FedAMP \cite{Huang_Chu_Zhou_Wang_Liu_Pei_Zhang_2021}, H-FedAMP \cite{Huang_Chu_Zhou_Wang_Liu_Pei_Zhang_2021}, pFedLA \cite{Ma_2022_CVPR} and FedALA \cite{Zhang_Hua_Wang_Song_Xue_Ma_Guan_2023} (personalized aggregation-based approach).

For each method, we save the best model that achieves the best performance on the local validation set during the training process, and evaluate it on the local test set of each client. The average of test accuracy across all clients is calculated.
All experiments are repeated for three times, and the results are reported in mean $\pm$ standard deviation.

\begin{table*}[htbp]
  \centering
  \small
    \begin{tabular}{lcccccccc}
    \toprule
    Settings & \multicolumn{4}{c}{Pathological setting} & \multicolumn{4}{c}{Practical setting (\textit{Dir}(0.1))} \\
    \midrule
     & CIFAR10 & CIFAR100 & $\text{CIFAR100}^*$ & TINY & CIFAR10 & CIFAR100 & $\text{CIFAR100}^*$ & TINY \\
    \midrule
    Local & 77.25±0.32 & 52.88±0.45 & 38.37±0.49 & 21.10±0.02 & 85.00±0.42 & 55.82±0.13 & 43.78±0.31 & 27.66±0.52 \\
    FedAvg & 66.03±0.84 & 36.03±0.36 & 19.50±0.43 & 10.83±0.14 & 60.72±1.13 & 35.48±0.07 & 18.84±0.62 & \multicolumn{1}{r}{9.79±0.18} \\
    FedAvg-FT & 83.46±0.26 & 60.88±0.23 & 48.20±0.39 & 34.29±0.29 & 87.54±0.24 & \underline{63.47±0.37} & \underline{52.70±0.23} & 38.19±0.18 \\
    FedProx & 66.35±0.84 & 36.16±0.38 & 19.89±0.44 & 10.87±0.24 & 60.58±0.84 & 35.60±0.20 & 18.78±0.08 & \multicolumn{1}{r}{9.66±0.28} \\
    FedProx-FT & 83.61±0.12 & \underline{60.96±0.01} & 48.33±0.40 & 34.40±0.34 & 87.65±0.34 & 63.26±0.36 & 51.89±0.21 & 38.57±0.19 \\
    \midrule
    Ditto & 82.80±0.57 & 59.73±0.25 & 46.26±0.12 & 33.44±0.09 & 86.77±0.27 & 61.98±0.40 & 50.62±0.27 & 37.26±0.50 \\
    FedRep & 82.14±0.12 & 55.44±0.56 & 42.95±0.62 & 31.03±0.50 & 87.27±0.09 & 58.80±0.41 & 47.96±0.63 & 35.25±0.49 \\
    pFedHN & 77.14±0.06 & 49.05±0.78 & 40.50±0.49 & 28.05±0.89 & 83.60±0.42 & 52.94±0.61 & 46.17±0.96 & 33.04±0.70 \\
    FedCP & 83.63±0.13 & 59.89±0.38 & 46.69±0.49 & 32.09±0.49 & 87.78±0.27 & 62.28±0.14 & 51.38±0.24 & 35.99±0.17 \\
    FedPAC & \textbf{83.74±0.15} & 60.14±0.67 & 47.69±0.88 & 33.26±0.40 & 87.59±0.17 & 62.32±0.10 & 51.80±0.06 & 37.15±0.29 \\
    \midrule
    CFL   & 81.89±0.36 & 58.36±0.35 & 46.59±0.31 & 34.46±0.42 & 87.17±0.16 & 61.11±0.16 & 51.14±0.27 & \underline{38.77±0.13} \\
    FedAMP & 77.29±0.04 & 52.92±0.35 & 37.93±0.40 & 21.61±0.06 & 85.07±0.43 & 56.52±0.07 & 43.81±0.82 & 27.75±0.10 \\
    H-FedAMP & 81.11±0.21 & 55.47±0.49 & 36.99±0.18 & \underline{34.79±0.27} & 86.01±0.32 & 58.24±0.72 & 42.22±0.33 & 38.49±0.16 \\
    pFedLA & 83.60±0.16 & 60.87±0.17 & \underline{48.76±0.49} & 34.19±0.11 & \underline{88.10±0.30} & 63.22±0.41 & 52.36±0.28 & 38.32±0.44 \\
    FedALA & 83.69±0.25 & 60.56±0.50 & 47.47±0.14 & 33.84±0.31 & 87.63±0.18 & 62.79±0.30 & 51.68±0.59 & 37.42±0.13 \\
    \midrule
    FedAGHN & \underline{83.72±0.18} & \textbf{61.27±0.14} & \textbf{48.92±0.80} & \textbf{34.97±0.38} & \textbf{88.32±0.29} & \textbf{63.66±0.22} & \textbf{52.75±0.37} & \textbf{38.96±0.37} \\
    \bottomrule
    \end{tabular}%
  \caption{The average test accuracy (\%) in pathological setting and practical setting on different datasets and models}\label{table:mainexp}
  \label{tab:addlabel}%
\end{table*}%

\subsection{Implementation Details}
In the specific experimental implementation, we use the SGD optimizer and set the local batch size to 64. The number of local training epochs is set to 5, and the local learning rate for models is set to 0.01. We consider 20 clients participating with the join rate of 1.
For the 4-layer CNN architecture with BN layers, we train it for 200 global communication rounds (1000 iterations). For ResNet-18, we train it for 80 rounds (400 iterations) on CIFAR100 and 40 rounds (200 iterations) on Tiny-Imagenet to ensure the empirical convergence. 

For the proposed FedAGHN, the learning rate of the hypernetwork is set to $\eta_{hn}= 0.005$. We tune $p_i$ from $\{0.02, 0.03, 0.06\}$ and $q_i$ from $\{0.1, 0.2, 0.5, 1, 3, 10\}$ for different datasets and Non-IID settings. 
For the other methods, we generally follow \cite{pmlr-v202-ye23b} to tune and set the hyper-parameters for these baselines\footnote{
For FedProx and FedProx-FT, we set the regularization weighted hyperparameter $\lambda = 0.01$. 
For Ditto, we use $\lambda = 1$ and $s=5$.
For FedRep, we set $\tau=5$. 
For pFedHN, we set $\eta_{hn}= 0.005$.
For FedCP, we set $\lambda=5$ on 4-layer CNN architecture with BN layers and $\lambda=1$ on ResNet-18.
For FedPAC, we set $\lambda=1$. 
For CFL, we set $\epsilon_1 = 0.5$, $\epsilon_2= 0.5$  on 4-layer CNN architecture with BN layers and $\epsilon_1 = 0.4$, $\epsilon_2= 1.6$ on ResNet-18, set $\text{minimum cluster size}=4$ on case study setting and $\text{minimum cluster size}=2$ on other settings, set $\text{start clustering round}=20/10/8$ for $\text{global rounds}=200/80/40$. 
For FedAMP, we set $\alpha_K =100$, $\lambda=1$, $\sigma=10000$. And for H-FedAMP, we set $\sigma=100$, $\omega_{ii}=0.05$, $\lambda=1$, $\alpha_K =1$ on 4-layer CNN architecture with BN layers and $\omega_{ii}=0.1$ on ResNet-18. 
For pFedLA, we set $\eta_{hn}= 0.005$, $\text{embedding dimension}=100$ and $\text{hidden layer dimension}=100$. For FedALA, we follow the setting as original paper.}.

\section{Results and Discussions}\label{sec:results}
To demonstrate the effectiveness of FedAGHN and its components, we investigate the following research questions.

\begin{itemize}
    \item RQ1: How does FedAGHN perform against the baseline methods in different scenarios? (Section \ref{sec:mainexp}) (Table \ref{table:mainexp})
    \item RQ2: Can and how the designs in FedAGHN contribute positively to performance? (Section \ref{sec:ablation} and Section \ref{sec:sensitivity}) (Table \ref{table:ablation} and Figure \ref{fig_sensitivity_pq})
    \item RQ3: Is FedAGHN still effective across various degrees of statistical heterogeneity as well as other scenarios? (Section \ref{sec:dir} and Section \ref{sec:other_scenarios}) (Table \ref{table:dir} and Table \ref{table:addnum})
    \item RQ4: Does the lightweight design of FedAGHN reduce the additional system overhead introduced by employing hypernetworks? (Section \ref{sec:systemoverhead}) (Table \ref{tab:overhead})
\end{itemize}

\subsection{Main Experiment}\label{sec:mainexp}

To verify the superiority of the proposed FedAGHN against the 15 baseline methods, we fairly test them over the aforementioned datasets and two different Non-IID settings. The best and second best performance are \textbf{bolded} and \underline{underlined}, respectively, in Table \ref{table:mainexp}.

Overall, FedAGHN outperforms almost all baselines and achieves the state-of-the-art performance for all cases across the two Non-IID scenarios. This might have been due to FedAGHN benefiting from its full consideration of personalization at the layer-wise level and its adaptively tuned client collaborative relationships.
The only case where FedAGHN does not achieve the first place is in the pathological setting of CIFAR10, where it takes the second place, narrowly trailing behind FedPAC. 
This may have been due to FedPAC making it easier to achieve feature alignment when both the number of model parameters and the variety of data sample labels are small. 
% The FedPAC introduces additional computation of feature centroids for feature alignment by regularization during local training, which allows better adaptation to local data but introduces additional communication costs with privacy concerns. 

The traditional FL methods FedAvg and FedProx perform worse than local training, whereas the local adaptation methods FedAvg-FT and FedProx-FT, remain competitive among various PFL approaches. 
%The competitive performances of FedAvg-FT and FedProx-FT are also mentioned in previous studies~\cite{xu2023personalized, pmlr-v202-ye23b}.

With regaed to the personalized aggregation-based methods, CFL performs competitively in the practical setting of TINY but performs worse in the other settings. This might have been due to the tendency of CFL to collaborate with a few clients whose updates of personalized local model parameters are most similar, while ignoring the general information from the other clients, making it less adaptable to different scenarios. 
On the other hand, H-FedAMP achieves high performance in the pathological setting of TINY and it outperforms FedAMP in most scenarios, possibly owing to its calculation of relative collaboration weights for each collaborating client, and its more flexible calculation of collaboration weights than that offered by FedAMP.

This study finds that finer-grained aggregation may result in improved performance. pFedLA performs well in most cases, possibly owing to its fine-grained personalized aggregation. However, pFedLA is not adaptable to different scenarios, possibly because of the reliance of its weight generation relies entirely on the optimization effects of the hypernetworks model parameters and the clients embeddings. FedALA performs worse than pFedLA in most cases, because it integrates information from both local and global models at a finer granularity but overlooks the selection of collaborative relationships with other clients. 
By comparison, our proposed FedAGHN outperforms all the personalized aggregation-based methods tested herein, possibly as a result of its ability to explicitly model collaborative relationships and adaptively tune collaboration weights while performing finer-grained aggregation.

\begin{table*}[htbp]
  \centering
  \small
    \begin{tabular}{lcccccccc}
    \toprule
    Design choices in FedAGHN & $p$  & $q$  & T & L & CIFAR10 & CIFAR100 & $\text{CIFAR100}^*$ & TINY \\
    \midrule
    FedAGHN w/o trainable parameter $p$ and $q$ & $\times$  & $\times$  & $\times$   & $\times$   & 88.17±0.13 & 63.34±0.53 & 52.19±0.13 & 38.33±0.54 \\
    FedAGHN w/ frozen trainable parameter $p$ and $q$ & \checkmark     & \checkmark     & $\times$   & $\times$ & 88.21±0.34 & 63.37±0.24 & 52.46±0.33 & \underline{38.92±0.31} \\
    FedAGHN w/o trainable parameter $p$ & $\times$   & \checkmark     & \checkmark     & \checkmark     & 88.20±0.16 & 63.45±0.21 & 52.18±0.41 & 38.62±0.03 \\
    FedAGHN w/o trainable parameter $q$ & \checkmark     & $\times$  & \checkmark     & \checkmark     & \underline{88.23±0.22} & \underline{63.57±0.08} & \underline{52.47±0.02} & 38.88±0.18 \\
    FedAGHN w/ the same $p$ and $q$ for all layers & \checkmark     & \checkmark     & \checkmark     & $\times$   & 88.07±0.26 & 63.47±0.08 & 52.14±0.25 & 38.89±0.21 \\
    \midrule
    FedAGHN & \checkmark     & \checkmark     & \checkmark     & \checkmark     & \textbf{88.32±0.29} & \textbf{63.66±0.22} & \textbf{52.75±0.37} & \textbf{38.96±0.37} \\
    \bottomrule
    \end{tabular}%
  \caption{The accuracy (\%) in practical heterogeneous setting for ablation study.}\label{table:ablation}
  \label{tab:addlabel}%
\end{table*}%

\subsection{Ablation Study}\label{sec:ablation}

To verify the effectiveness of the designed components in FedAGHN, ablation experiments are performed in the practical setting. We consider trainable parameters $p$ and $q$ in AGHN, together with their trainability (T) and layer-wise differences (L). 
The results of the ablation experiments are shown in Table \ref{table:ablation}, where `with' is abbreviated as `w/' and `without' is abbreviated as `w/o'. 

FedAGHN performs best when all of its designed components are fully considered, reflecting the positive contribution of each of the designed components to the overall performance.
Specifically, first of all, the most severe performance decrease occurs when $p$ and $q$ are absent, highlighting the significance of FedAGHN in explicitly tuning collaborative relationships. Second, freezing $p$ and $q$ during the FL process leads to performance decrease, which demonstrates that AGHN's adaptive optimization of trainable parameters can better tune current collaborative relationships. Third, removing $p$ and $q$ separately results in a greater decrease in performance without $p$ than without $q$, indicating the importance of tuning the client's self-weighting value and adaptive optimization. Finally, performance also decreases when there is no layer-wise difference in $p$ and $q$, highlighting the importance of layer-wise collaboration tuning.

\subsection{Different Degrees of Heterogeneity}\label{sec:dir}

\begin{figure*}[htbp]
\centering
\includegraphics[width=0.99\textwidth]{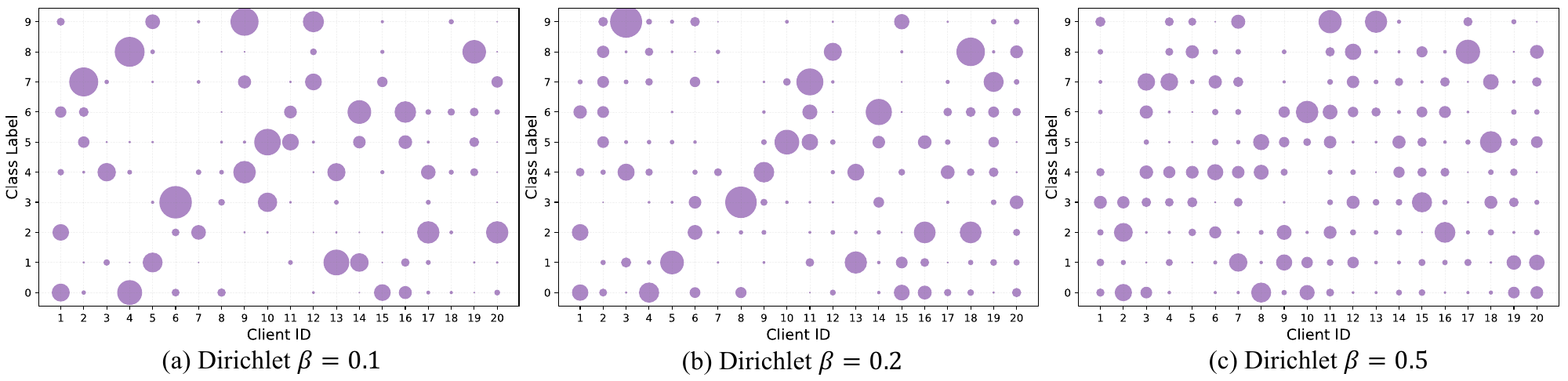}
\caption{Visualization of data distribution with different degrees of statistical heterogeneity among 20 clients on CIFAR10, where the size of dots denotes the number of samples per class distributed to each client.
}
\label{fig_dir_distribution}
\end{figure*}

To investigate the performance of FedAGHN under various degrees of data heterogeneity, we manipulate the parameter $\beta$ in \textit{Dir}($\beta$) to generate heterogeneous data with different distributions on the CIFAR10 dataset. A smaller $\beta$ indicates higher data heterogeneity as illustrated in Figure \ref{fig_dir_distribution}. 
The experimental results of FedAGHN and the baseline methods under various degrees of data heterogeneity are presented in Table~\ref{table:dir}.

\begin{table}[ht!]
  \centering
  \small
    \begin{tabular}{lccc}
    \toprule
    Settings & \textit{Dir}(0.1) & \textit{Dir}(0.2) & \textit{Dir}(0.5) \\
    \midrule
    Local & 85.00±0.42 & 80.47±0.21 & 71.41±0.12 \\
    FedAvg & 60.72±1.13 & 60.04±1.52 & 65.06±0.11 \\
    FedAvg-FT & 87.54±0.24 & 84.95±0.34 & 79.51±0.16 \\
    FedProx & 60.58±0.84 & 60.02±1.28 & 65.05±0.08 \\
    FedProx-FT & 87.65±0.34 & 84.69±0.33 & 79.60±0.33 \\
    \midrule
    Ditto & 86.77±0.27 & 84.19±0.32 & 79.11±0.43 \\
    FedRep & 87.27±0.09 & 84.22±0.31 & 77.47±0.34 \\
    pFedHN & 83.60±0.42 & 80.83±0.34 & 73.18±0.41 \\
    FedCP & 87.78±0.27 & 85.25±0.42 & 79.57±0.16 \\
    FedPAC & 87.59±0.17 & \underline{85.39±0.10} & 79.35±0.66 \\
    \midrule
    CFL   & 87.17±0.16 & 83.96±0.18 & 77.13±0.25 \\
    FedAMP & 85.07±0.43 & 80.57±0.33 & 71.77±0.23 \\
    H-FedAMP & 86.01±0.32 & 82.64±0.64 & 75.77±0.55 \\
    pFedLA & \underline{88.10±0.30} & 84.97±0.16 & 79.37±0.56 \\
    FedALA & 87.63±0.18 & 84.89±0.21 & \underline{79.64±0.36} \\
    \midrule
    FedAGHN & \textbf{88.32±0.29} & \textbf{85.45±0.29} & \textbf{79.85±0.52} \\
    \bottomrule
    \end{tabular}%
  \caption{The accuracy (\%) on \text{CIFAR10} for different heterogeneity.}
  \label{table:dir}
\end{table}%

FedAGHN outperforms all baselines across various degrees of data heterogeneity, which might have been owed to the consideration of fine-grained aggregation and personalization at layer-wise. It is evident that some of the personalized aggregation-based methods, such as CFL, FedAMP and H-FedAMP, yield show lower performance, whereas the fine-grained personalized aggregation-based methods such as pFedLA and FedALA perform well. 
Furthermore, pFedLA and FedPAC show competitive performance, possibly as a result of their capability to partially address the personalization needs of different model layers.

\subsection{Performances on Other Scenarios}\label{sec:other_scenarios}

In addition, we have conducted the following experiments to account for a broader range of realistic scenarios.

\textit{Extreme heterogeneity:} To simulates a scenario of extreme heterogeneity in label distribution, we divide the CIFAR-100 dataset into 20 independent clients, each client is randomly assigned local data from 5 distinct labels, with no label overlap across clients (i.e., each client has an exclusive set of 5 labels, ensuring completely disjoint label distributions).

\textit{Cross-device FL:} To verify the performance under the cross-device FL scenario with more client participation, we expand the number of clients to 50/100 in the pathological heterogeneous scenario of the CIFAR-100 dataset, while maintaining the other experimental settings consistent with the main experiment. To ensure that the amount of data in each category for each client is consistent, we assign 20/8/4 classes to each client when the number of clients is 20/50/100, respectively.

The results for these two additional scenarios are summarized as Table \ref{table:addnum}. The experimental results show that our method can still maintain the superior performance under this extremely heterogeneous scenario and with an even larger number of client participation, consistent with the existing experimental conclusions.

\begin{table}[htbp]
  \centering
  \small
  \setlength{\tabcolsep}{3pt}
    \begin{tabular}{lcccc}
    \toprule
    
    Datasets & Extreme & Num=20 & Num=50 & Num=100 \\
    \midrule
    Local & 81.53±0.44  & 52.88±0.45 & 66.98±0.53 & 78.00±0.10 \\
    CFL   & 83.30±0.07 & 58.36±0.35 & 70.38±0.34  & 80.21±0.06  \\
    H-FedAMP & 80.43±0.10  & 55.47±0.49 & 66.79±0.49 & 77.41±0.30 \\
    FedAGHN & 83.46±0.22 & 61.27±0.14 & 73.63±0.17  & 80.69±0.33 \\
    \bottomrule
    \end{tabular}%
  \caption{The accuracy (\%) of compared methods in case study}
  \label{table:addnum}
\end{table}%
\subsection{Sensitivity Analysis on p and q}\label{sec:sensitivity}

For the proposed FedAGHN, for each client i, we tune $p_i$ from $\{0.02,0.03,0.06\}$ and $q_i$ from $\{0.1,0.2,0.5,1,3,10\}$ across different datasets and Non-IID settings. The initial values of $p_i$ and $q_i$ are applied uniformly to all R layers of the model.

\begin{figure}[ht!]
\centering
\includegraphics[width=0.49\textwidth]{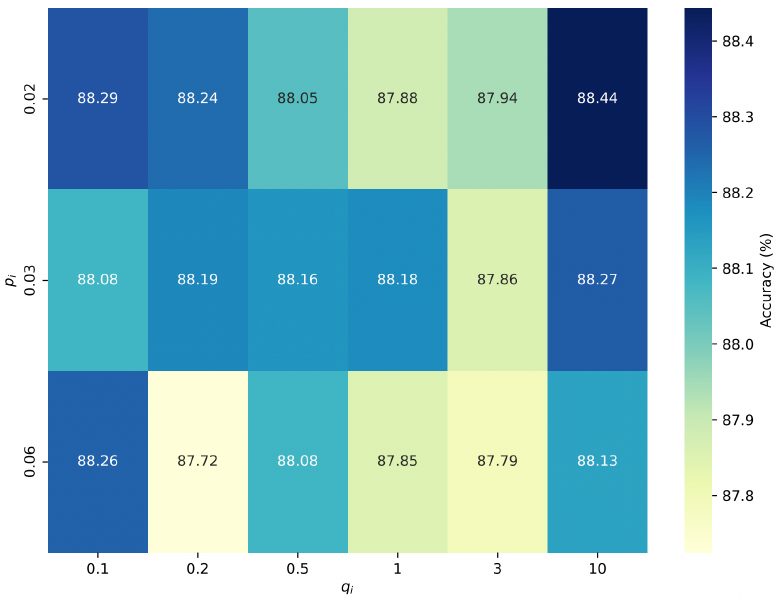}
\caption{Sensitivity analysis on p and q using the CIFAR-10 dataset under the Dir(0.1) partition setting.}
\label{fig_sensitivity_pq}
\end{figure}

To further explore the roles of $p_i$ and $q_i$ in our method, we conduct a sensitivity analysis on $p_i$ and $q_i$ using CIFAR-10 under the Dir(0.1) partition setting. The results are presented in Figure \ref{fig_sensitivity_pq}. The selections of $p_i$ and $q_i$ are as following:

\begin{enumerate}
    \item $p_i$ represents the self-participation strength of client i. When the number of clients is 20, the average weight is approximately 0.05. Therefore, we set the initial range of $p_i$ to $\{0.02,0.03,0.06\}$ to reflect different degrees of self-participation.
    \item $q_i$ serves as a scaling coefficient that adjusts the variability of collaboration weights (which can also be interpreted as the reciprocal of a temperature coefficient). Accordingly, $q_i$ is tuned from $\{0.1,0.2,0.5,1,3,10\}$ to capture varying levels of weight differentiation (or smoothness).
\end{enumerate}

Overall, FedAGHN exhibits low sensitivity to these parameters. While different combinations of $p_i$ and $q_i$ introduce some variation in performance, the accuracy difference remains within approximately 0.8\%. This robustness can be attributed to the dynamic optimization strategy and the integration of prior knowledge within AGHN’s tunable attention mechanism. Although the selection of $p_i$ and $q_i$ in the tunable attention mechanism introduces some variability, the dynamic optimization and prior knowledge substantially enhance the overall robustness.

\subsection{System Overhead}\label{sec:systemoverhead}

To assess the effectiveness of FedAGHN's lightweight design for hypernetworks, Table \ref{tab:overhead} presents a comparative analysis of the system overhead incurred by PFL methods employing hypernetworks.
% We analyze the system overhead from three perspectives and the domain shift scenario as summarized in the table.

\textbf{Communication:} Each client incurs a communication cost of $S$ for transmitting its local model. 
All PFL methods employing hypernetworks involve no additional communication overhead, as the hypernetworks are maintained on the server side.
%Except for FedPAC, which additionally transmits feature centroids, all other methods involve no extra communication overhead.

\textbf{Storage:} For server-side storage cost (maximum required space during computation), $M$ denotes the local model parameter size, with $N$ clients participating. $M_{\text{hn}}$ represents storage overhead for maintaining hypernetwork. $M_{\text{hn}}$ varies across algorithms (pFedHN, pFedLA and FedAGHN for 1.13B, 115K and 82 trainable parameters). FedAGHN, which focuses on the design of critical parameters, has the smallest hypernetwork size. Thus, the model size $M$ is 11.23M, and the storage cost of $M_{\text{hn}}$ in FedAGHN is much smaller than that of $M$.

\textbf{Computation:} In terms of overall computation overhead and time during the aggregation phase, FedAGHN shows relatively low cost compared to other personalized aggregation methods, despite maintaining $N$ AGHNs. This efficiency benefits from the introduction of a small number of key parameters in the AGHN's lightweight attention mechanism.
$C$ denotes the computational cost of the weighted aggregation operation, and $C_{\text{hn}}$ represent the computational overhead of performing personalized aggregation using the hypernetwork. Set $N=20$ to evaluate time costs.
The computation cost $C_{\text{hn}}$ of AGHN comprises three components: attention computation, on-graph aggregation, and hypernetwork update. The computation times for pFedHN, pFedLA, and FedAGHN are 1.77s, 1.54s, and 1.02s, respectively.

\textbf{Scalability:} For the scaling overhead, although FedAGHN maintains $N\times R$ collaboration graphs, its primary purpose is to enable personalized aggregation. Once the personalized aggregation is completed, the corresponding collaboration graph no longer needs to reside in memory (regardless of whether it is at the layer-wise or the client-wise). For example, after the $r$-th layer of client $i$ has completed personalized aggregation, $G_i^r$ can be safely discarded. Consequently, even when the number of clients $N$ scales to a very large value, this design prevents excessive memory overhead.

\begin{table}[htbp]
  \centering
  \small
  \setlength{\tabcolsep}{4pt}
  %\caption{Add caption}
    \begin{tabular}{lccc}
    \toprule
    Algorithm    & Comm. & Server Storage & Computation Cost \\
    \midrule
    pFedHN & $S$     & $ M + M_{\text{hn}}$ & $C_{\text{hn}}$ \\
    pFedLA & $S$     & $N \times M + N \times M_{\text{hn}}$ & $N \times C_{\text{hn}}$ \\
    FedAGHN & $S$     & $N \times M + N \times M_{\text{hn}}$ & $N \times C_{\text{hn}}$ \\
    \bottomrule
    \end{tabular}%
  \caption{Comparison of system overhead for PFL methods employing hypernetworks}
  \label{tab:overhead}%
\end{table}%

\section{Analysis of Collaboration in FedAGHN}\label{sec:collaboration}
In addition to the above experiments, we investigate the following research questions to further explore the collaborative relationship learned by FedAGHN.

\begin{figure*}[htbp]
\centering
\includegraphics[width=0.9\textwidth]{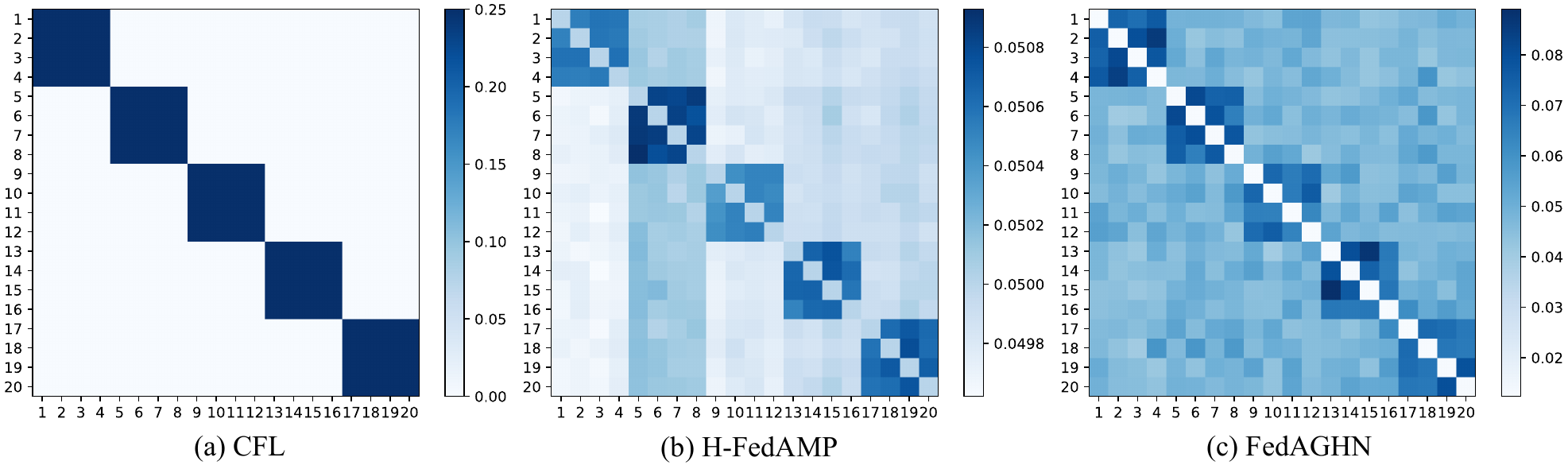}
\caption{Visualization of collaboration graphs under the case study setting on CIFAR100. (a) and (b) show the collaboration graphs of CFL and H-FedAMP respectively, which hold a same collaboration graph for all layers. (c) FedAGHN shows the average graph over all layers.}
\label{fig_visual_compare}
\end{figure*}

\begin{itemize}
    \item RQ5: Can the collaboration graphs in FedAGHN effectively capture the collaborative relationships among clients? (Section \ref{sec:visual_compare}) (Figure \ref{fig_visual_compare}, Table \ref{table:case}) % \ref{table:caseweight}
    \item RQ6: Can FedAGHN meet the personalized requirements for layers and show differences among different layers? (Section \ref{sec:visual_layer}) (Figure \ref{fig_visual_layer})
    \item RQ7: Can FedAGHN dynamically tune collaborative relationships during FL process? (Section \ref{sec:visual_rounds}) (Figure \ref{fig_visual_round})
    \item RQ8: How do the collaborative relationships of different model layers captured by FedAGHN change throughout the FL process? (Section \ref{sec:visual_rounds}) (Table~\ref{table:rounds})
\end{itemize}

\subsection{Case Study Setting}

In this section, we establish a case study setting to investigate the collaborative relationship learned by FedAGHN. The Non-IID scenario is introduced following~\cite{xu2023personalized}. 
We randomly sample $s$\% of the data (80\% by default) and divide them into 5 groups based on the classes of labels. The $N=20$ clients are divided into 5 groups, and each group is assigned data from the same classes. The remaining 1-$s$\% of the data is distributed to all clients in an IID manner. 

Therefore, clients within each group follow the same data distribution, while clients across groups have different dominant classes. In this context, there are evident differences between clients within the same group and clients in different groups, and thus, we can intuitively estimate whether the collaborative relationships among clients are appropriately captured.

%in the case study setting
Table \ref{table:case} has shown the performance of these methods in case study setting.
FedAGHN performs best on these datasets in this setting, which could be explained by the appropriate collaboration graphs analyzed in Section \ref{sec:analysisgraph} and \ref{sec:visual_rounds}.
CFL has relatively high performance in most cases, possibly because CFL can collaborate with the clients in the same group, but it does not collaborate with the clients in different groups at all, ignoring the information from clients with different data distributions. H-FedAMP assigns smaller weights to the clients in different groups, but its performance is even worse than local training on $\text{CIFAR100}^*$, probably because the differences in weights among clients may not be sufficiently large.

\begin{table}[htbp]
  \centering
  \small
  \setlength{\tabcolsep}{3pt}
    \begin{tabular}{lcccc}
    \toprule
    Datasets & CIFAR10 & CIFAR100 & $\text{CIFAR100}^*$ & TINY \\
    \midrule
    Local & 73.96±0.10 & 40.25±0.25 & 28.56±0.40 & 14.99±0.04 \\
    CFL   & \underline{79.66±0.21} & \underline{49.55±0.34} & \underline{38.59±0.46} & 25.93±0.29 \\
    H-FedAMP & 78.23±0.49 & 46.11±0.67 & 27.09±0.81 & \underline{26.74±0.22} \\
    FedAGHN & \textbf{81.09±0.30} & \textbf{51.14±0.27} & \textbf{39.12±0.48} & \textbf{26.83±0.14} \\
    \bottomrule
    \end{tabular}%
  \caption{The accuracy (\%) of compared methods in case study}
  \label{table:case}
\end{table}%

\subsection{Analysis of Collaboration Graphs}\label{sec:analysisgraph}
\subsubsection{Comparison with other methods}\label{sec:visual_compare}

To evaluate the efficacy of personalized aggregation methods in capturing collaborative relationships, the visualization of collaboration graphs in CFL, H-FedAMP and FedAGHN are shown in Figure~\ref{fig_visual_compare}-a, \ref{fig_visual_compare}-b and \ref{fig_visual_compare}-c respectively.

As illustrated in Figure~\ref{fig_visual_compare}, FedAGHN can effectively capture collaborative relationships among clients, with small self-weighting value. Overall, the collaboration weights of FedAGHN generally align with the similarity of clients' data distributions in the case study setting. Moreover, FedAGHN has clear weight differences both intra-group and inter-group.
For the clients within a same group, there are significant weight differences among clients, which implies that higher weights are assigned to more beneficial clients.
Besides, it is shown in Figure \ref{fig_visual_compare}-a that all clients within a group are assigned the same collaboration weights in CFL, while clients outside the group do not participate in collaboration. Furthermore, H-FedAMP introduces a fixed self-weighting value, while assigning slightly higher weights to clients within the same group and lower weights to others.

To further analyze the collaboration weights among clients with different relationships, the averaged collaboration weights of $\textit{N}$ clients are shown in Table \ref{table:caseweight}. The collaboration weights among clients with different relationships are divided into three types, where the self-weighting value of central client is denoted as `Self', the average weight of the other clients in the same group with central client is denoted as `Similar' and the average weight of clients in the groups different from the central client is denoted as `Other'.

\begin{table}[b]
  \setlength{\tabcolsep}{11pt}
  \centering
  \small
  {
  \begin{threeparttable}
    \begin{tabular}{lccc}
    \toprule
    Methods & Self  & Similar & Other \\
    \midrule
    CFL   & 0.2500  & 0.2500  & 0.0000  \\
    H-FedAMP & 0.0500  & 0.0506  & 0.0499  \\
    FedAGHN & 0.0126  & 0.0733  & 0.0480  \\
    \bottomrule
    \end{tabular}%
    \end{threeparttable}
    }
  \caption{Average collaboration weights of different collaborating clients on CIFAR100.} \label{table:caseweight}
  \label{tab:addlabel}%
\end{table}%

In FedAGHN, the averaged collaboration weights are significantly different for clients with different relationships, where the self-weighting value is smaller, whereas the weights for clients within the same group are larger. This might be because of the greater diversity of information among clients in the same group. On the other hand, CFL maintains the same weight for clients within a group, whereas H-FedAMP assigns slightly different weights for clients with different relationships.

\subsubsection{Layer-wise collaboration graphs}\label{sec:visual_layer}

\begin{figure}[t]
\centering
\includegraphics[width=0.48\textwidth]{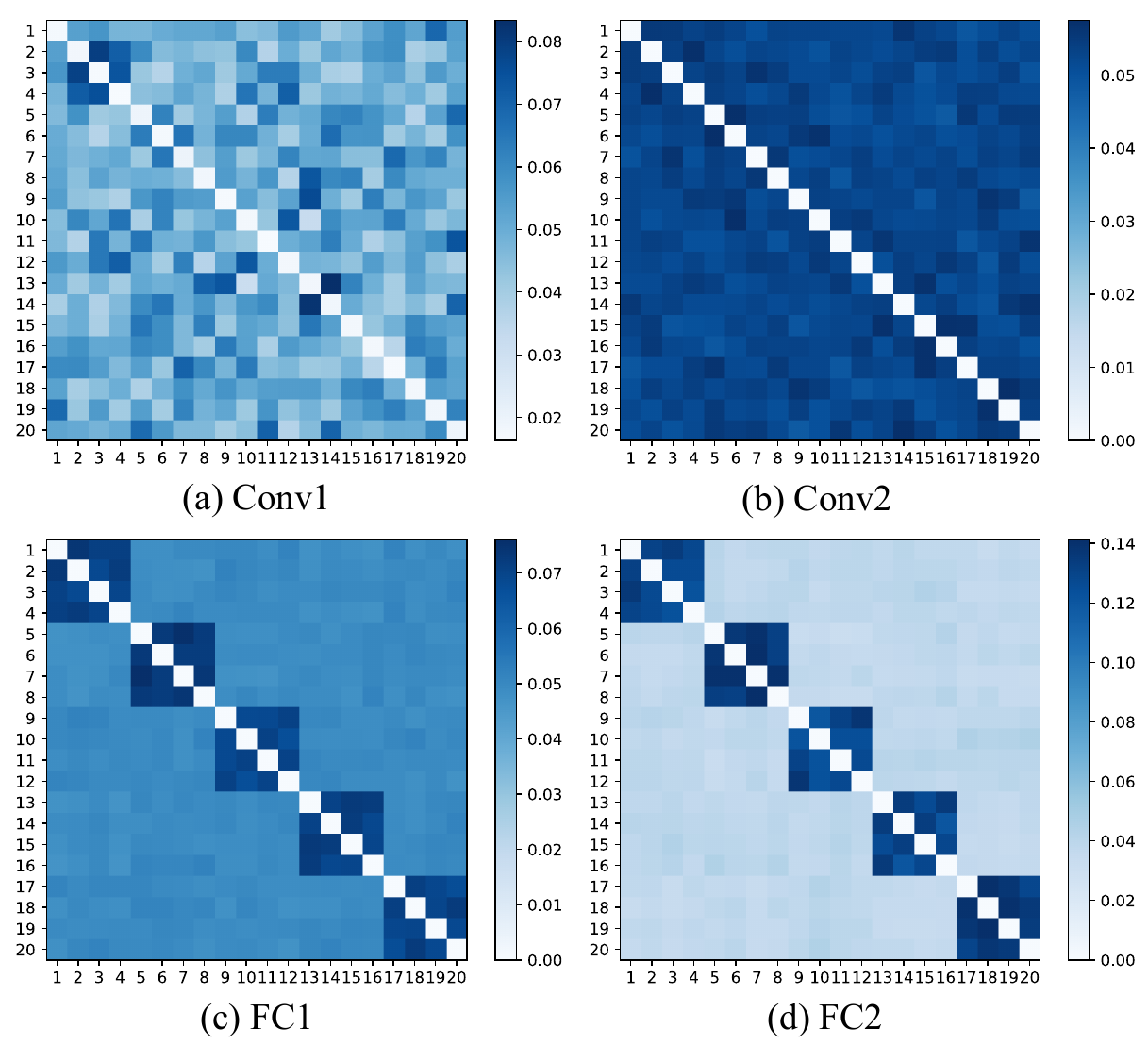}
\caption{Visualization of the FedAGHN's collaboration graphs for different model layers in 4-layer CNN with BN layers model under case study setting on CIFAR100.}
\label{fig_visual_layer}
\end{figure}

To further investigate layer-wise personalization in FedAGHN, we visualize the collaboration graphs corresponding to different model layers on CIFAR100. Figure~\ref{fig_visual_layer} reveals significant differences in the collaborative relationships captured by the shallow and deep layers of FedAGHN.

In the shallow layers (Figure~\ref{fig_visual_layer}-a, \ref{fig_visual_layer}-b), the model tends to select more general information, with no significant differences between groups. 
FedAGHN introduces the tunable attentive mechanism to capture the fine-grained collaborative relationships among clients, which enables it to focus more on valuable information in the shallow layers, without excluding clients with different data distributions. 

In the deep layers (Figure~\ref{fig_visual_layer}-c,~\ref{fig_visual_layer}-d), the model tends to select more specific information and prefers collaboration among clients within the same group. Significant differences can be observed between the groups, especially in the last model layer.

Additionally, we find that setting the self-weighting value to a relatively smaller value leads to better performance in our experiments. The reason behind this could be that FedAGHN provides the detailed descriptions of collaborative relationships. Under the assumption that the collaborative relationships are captured effectively, reducing the self-weighting value facilitates the acquisition of more accurate and diverse information.

\subsection{Analysis of Dynamic Collaboration}\label{sec:visual_rounds}

\begin{figure*}[hb!]
\centering
\includegraphics[width=0.99\textwidth]{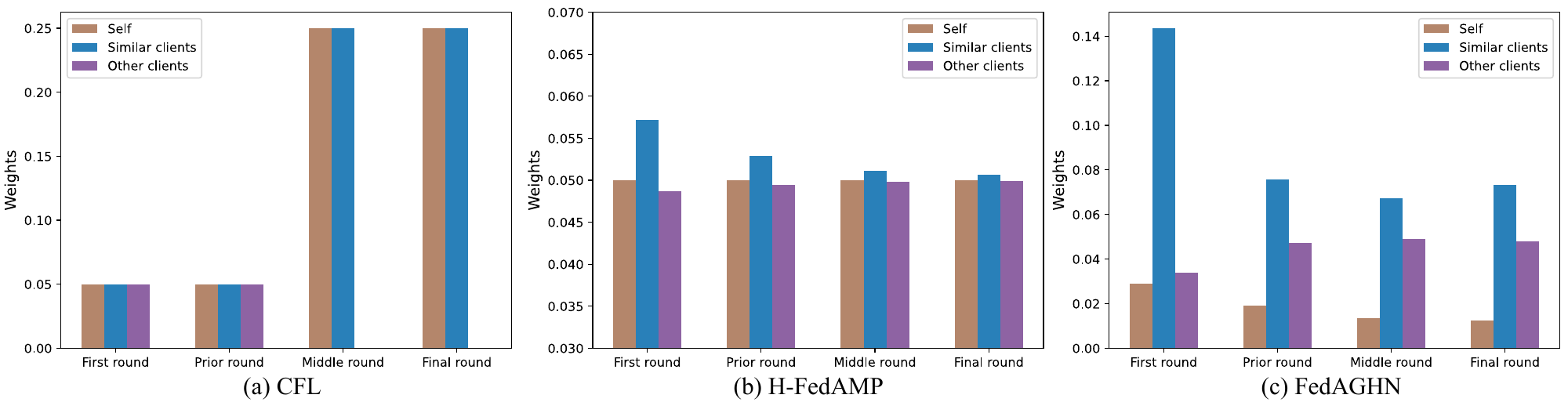}
\caption{The evolution trends of collaboration weights for compared methods at various stages, which are counted under case study setting on CIFAR100. 
The final round is the round in which we report the best-performing model. The prior round and the middle round are set as the 1/6 and 1/2 of the final round, respectively. 
}
\label{fig_visual_round}
\end{figure*}

To demonstrate that the proposed FedAGHN can dynamically tune the client collaboration during the training process, we record the changes of collaboration weights for different methods during the first, prior, middle, and final round of training. 
Figure \ref{fig_visual_round} illustrates the evolution of collaboration weights among different methods.

Overall, the weight differences among collaborating clients continue to change during the training process, possibly indicating that the adaptive tuning of collaborative relationships may enhance FedAGHN's performance. FedAGHN has a high demand for collaboration among clients with similar data distributions in the first round, which stabilizes in the middle round, but significant differences still exist among clients with different collaboration relationships. 
The self-weighting value gradually decreases during the FL process, indicating that clients prefer to collaborate with others. Therefore, FedAGHN's adaptive tuning of collaborative relationships can better fulfill the collaboration requirements at different stages.

For comparison, the evolution of collaboration weights for CFL is dependent on clustering, where similar clients are given high weights but the collaboration with other clients is completely ignored. Meanwhile, the weight differences among the collaborating clients in H-FedAMP are slight, and the self-weighting value remains constant throughout the training process.

To analyze the evolution trends of the collaborative relationships for different model layers, Table \ref{table:rounds} shows the changes of weights for the whole model, shallow layers and deep layers of model in FedAGHN. 
The definition of different stages follows the description in Figure \ref{fig_visual_round} and the definition of `Self', `Similar' and `Other' follows the explanation in Section \ref{sec:visual_compare}.
It can be observed that FedAGHN fully considers the differences in collaborative requirements across different model layers.

\begin{table}[t]
  \centering
  \small
  \setlength{\tabcolsep}{4.5pt}
    \begin{tabular}{llcccc}
    \toprule
          & Client & First & Prior & Middle & Final \\
    \midrule
    $\text{FedAGHN}_\text{all}$ & Self  & 0.0291  & 0.0192  & 0.0136  & 0.0126  \\
    $\text{FedAGHN}_\text{all}$ & Similar & 0.1436  & 0.0758  & 0.0674  & 0.0733  \\
    $\text{FedAGHN}_\text{all}$ & Other & 0.0338  & 0.0471  & 0.0490  & 0.0480  \\
    \midrule
    $\text{FedAGHN}_\text{shallow}$ & Self  & 0.0291  & 0.0228  & 0.0175  & 0.0155  \\
    $\text{FedAGHN}_\text{shallow}$ & Similar & 0.1270  & 0.0703  & 0.0598  & 0.0586  \\
    $\text{FedAGHN}_\text{shallow}$ & Other & 0.0369  & 0.0479  & 0.0502  & 0.0505  \\
    \midrule
    $\text{FedAGHN}_\text{deep}$ & Self  & 0.0291  & 0.0155  & 0.0097  & 0.0097  \\
    $\text{FedAGHN}_\text{deep}$ & Similar & 0.1601  & 0.0814  & 0.0750  & 0.0880  \\
    $\text{FedAGHN}_\text{deep}$ & Other & 0.0307  & 0.0463  & 0.0478  & 0.0454  \\
    \bottomrule
    \end{tabular}%
  \caption{Average collaboration weights of different collaborating clients for FedAGHN's layers at certain rounds on CIFAR100.}\label{table:rounds}
  \label{table}%
\end{table}%

In the shallow layers, the weight differences among collaborating clients gradually decrease, and in the final round, the difference between similar clients and other clients is only 0.008, which aligns with the model's requirement to capture more generic information in shallow layers. In the deep layers, the weight differences among collaborating clients remain significantly larger, and the average collaboration weight of clients with similar data distributions in the final round is approximately twice that of the clients with different data distributions, consistent with the model's requirement for more specific information in deep layers.

\section{Conclusion}
This work proposed FedAGHN, a personalized aggregation-based method, to continuously capture appropriate collaborative relationships for fine-grained personalized aggregation. In each communication round, FedAGHN maintains a set of collaboration graphs for each client and each layer, which explicitly model the client-specific and layer-wise collaboration weights. The collaboration weights are obtained by the newly proposed tunable attentive mechanism. Finally, the personalized initial models are obtained via aggregating parameters over the learned collaboration graphs. 
Extensive experiments demonstrated the superiority of FedAGHN, which is attributed to its three important innovations (Layer-wise collaboration graphs, Attention-based mechanism with trainable parameters, and Personalized aggregation guided by collaboration graphs).
We also performed experiments to analyze how FedAGHN uniquely captures collaborative relationships across different model layers, various stages of the FL process, and self-participation weight.

% For future work, we suggest further understanding the evolution of collaborative relationships during the FL process, not only for FedAGHN but also for other PFL methods. 
For future work, we suggest that whether there are more suitable input features is a problem worth exploring. In addition, the ability of capturing collaboration graphs is limited by the capacity of the hypernetwork, and the complexity of the hypernetwork and the learning cost would be a balance point worth careful study. Finally, FedAGHN is an aggregation optimization method that is decoupled from the optimization of the local update process, and combining it with other local update optimization methods is also a potential extension point.

% \newpage

\bibliographystyle{named}
\bibliography{ijcai25}

\end{document}